# Abnormal Clients Clustering on Federated Learning Model

**First A. Jihwan Won[1]**
[1]School of Computer Science, Kwangwoon University.

Corresponding author: First A. Author (e-mail: wjh5597486@ gmail.com).

**ABSTRACT** Federated learning is a model for privacy without revealing private data by transfer models instead of personal and private data from local client devices. While, in the global model, it's crucial to recognize each local data is normal. This paper suggests one method to separate normal locals and abnormal locals by Euclidean similarity clustering of vectors extracted by inputting dummy data in local models. In a federated classification model, this method divided locals into normal and abnormal.

**INDEX TERMS** Federated learning, Sybil attack.

## I. INTRODUCTION

Federated Learning has been issued as an extraordinary model for preserving private data. It makes the machine learning model in the global model train without sharing personal and private data which is distributed to many other local devices. However, it is also unable to access the data and analyze it to tell which local data is poisoned[1]. With this disadvantage, Federated learning is easily exposed to Sybil attacks such as transfers of faulty results to the global model. FoolsGold[2], those key is that Sybil's training directions are more similar to other Sybil's than typical average local's directions, is known as one of the best algorithms for defense from Sybil attacks. Likewise, FoolsGold algorithm, This paper shows a similarity-based algorithm that detects Sybils with a classification of vectors from each local model by inputting dummy data.

## II. BACKGROUND

### A. Federated Learning Model
In Federated learning training, each local client gets the ML model from the global server and transfers the model after training with data in the locals. The global server gathers all the models from local clients and updates the global model. the updated model is transferred to each local. The representative method of updating is FedAvg[3] which naively computes the model by averaging for all model's weights.

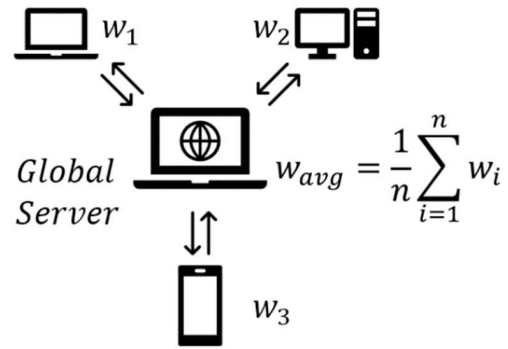

**FIGURE 1.** Federated Averaging Architecture

### B. KNN Clustering
K-nearest Neighbor(KNN) Algorithm is a supervised learning algorithm for classification. KNN gives a new data label by k number of points that locate near the data point. In high dimensional Euclidean Distance (eq. 1) is one of the most common ways to determine the distance in high dimensional spaces for the difference between them.

$$d(p,q) = \sqrt{\sum_{i=1}^{n}(q_i - p_i)^2}$$

(1)

## C. Poisoning Attack

In many fields, Machine Learning requires numerous data including personal information. Such as personal health, wealth, and address. Thus, it's necessary to train machine learning models without expelling data from the device. Federated learning is one of the finest methods to train models with keeping the private. However, the server that has the global model can't deal with the data. The server receives only privately trained models for updating the global model. It makes it easy to attack the global model by transferring the wrong model. For instance, Sybils can train the model with label-flipped data which is labeled wrong or add some noise to the dataset.

## III. ALGORITHM

### A. Assumption

This work is based on the following Assumptions.
1) All Attackers cannot attack the global model directly, they only transfer maliciously trained models.
2) Each attacker has its own data. It is unable to access other honest clients' data.
3) We can make a hypothesis that the normal model's output is more similar to other normal models' output than the abnormal models' output. As train directions are similar to others among the normal models[2]. Thus, outputs from honest models and attackers are extinguishable by comparing Euclidean Distances.

### B. Proposed Algorithm

Indicate locals are put into the locals. Half of them have a normal dataset, while the others have a noisy dataset. These make both Honest and Sybil class indicators in KNN clustering. After each round of training, we put 28x28 noisy data into local models. As a result, we get 10-length vectors from each of the client models. And then, apply KNN clustering. Those local models filtered by KNN clustering are excepted from the update for the round to prevent negative affecting on the global model. In this work, we set k=1.

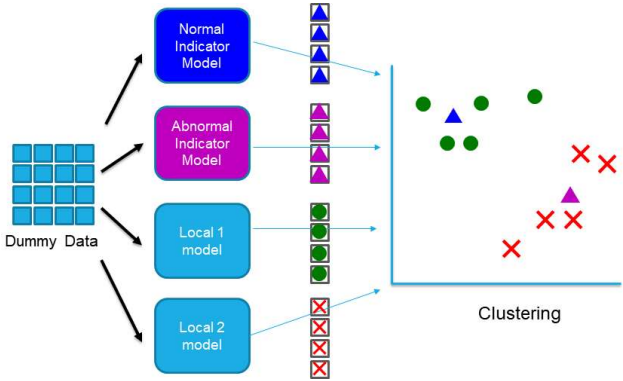

**FIGURE 1.** Process of Clustering of Vectors

## IV. EXPERIMENT

### A. DATASET

MNIST is utilized for this experiment. The dataset consists of 60,000 of 28x28 grayscale data. each has one class out of 10 classes. Each local has its dataset without overlap data.
Totally 100 clients are set for training, each has 600 MNIST IID data. Two of them are normal indicator clients which have normal data and the other two having noisy data are named abnormal indicators. 48 normal and 48 abnormal clients are filtered by the proposed algorithm.

TABLE I
DATASET OF LOCALS

| Category 1 | Category 2 | Dataset | Local Number |
|---|---|---|---|
| Indicator Client | Normal | Normal Data | 2 |
|  | Abnormal | Random Noisy Data | 2 |
| Subject Client | Normal | Normal Data | 48 |
|  | Abnormal | Random Noisy Data | 48 |

### B. FEDERATED LEARNING CLIENT SETTING

The global server and the locals have the same CNN architecture model which is described in Table 2. Learning-rate is 0.001. Adam is utilized as the optimizer. All clients train 5 epochs per one round. During 30 rounds, the global updates when each round is finished.

TABLE 2
CNN MODEL ARCHITECTURE

| Layer | Output size |
|---|---|
| Input data | 1@28x28 |
| 3x3 convolution | 16@28x28 |
| ReLU | - |
| 2x2 Max-Pooling | 1 |
| 3x3 convolution | 16@14x14 |
| ReLU | 32@14x14 |
| 2x2 Max-pooling | 32@7x7 |
| Flatten | 1568 |
| Fully Connected | 256 |
| ReLU | - |
| Fully Connected | 10 |

|  | Soft Max | - |

## D. RESULT

the performance of the algorithm is evaluated after 30-round training. The results are shown in Figure 2. There are 10 cases of noise scale from 0 to 1. 0-ratio stands for Sybil locals have normal data while 1-ratio stands for Sybil locals are only random noise. In this case, the ratio is 0.5, and each data is the aggregation result of 50 % random noise and 50 % normal data. the accuracy goes down with the noise rate until the rate is 0.6.

Even though the proposed algorithm cannot filter perfectly in small-scale cases, it shows better results than the situations exposed by Sybils. And the proposed algorithm can tell easier the malicious clients which have the stronger noise.

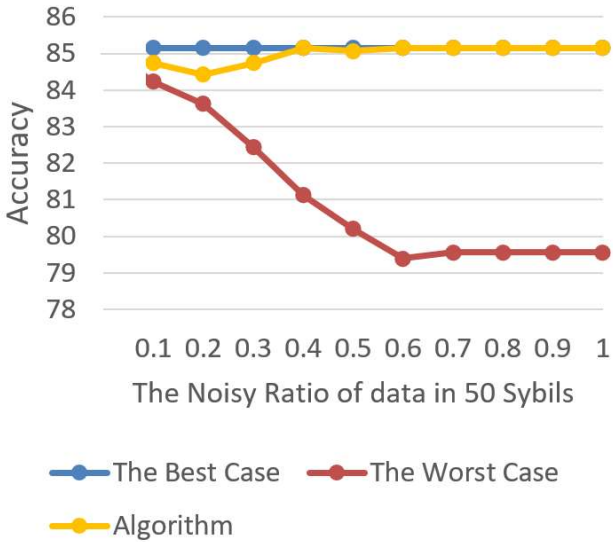

**FIGURE 2.** Accuracy for Validation Dataset. The X-axis stands for the ratio of the noise in image data in 50 abnormal clients. The best-case accuracy is the result that we get when the filter perfectly separates normal and abnormal clients. The worst case is when the filter cannot tell abnormal clients, so the global model is trained by all normal and abnormal clients.

Figure3 shows how the vectors can be clustered by Euclidean distance at round 0 and round 30. And Figure 4 denotes that the number of abnormal clients doesn't affect the performance of the algorithm. Moreover, the more noise it has, the more similar it is to each other among Sybils. Noisy locals locate distinctly from the Sybil clients having mistakenly labeled data. With these results, we can presume the attack method of Sybil clients.

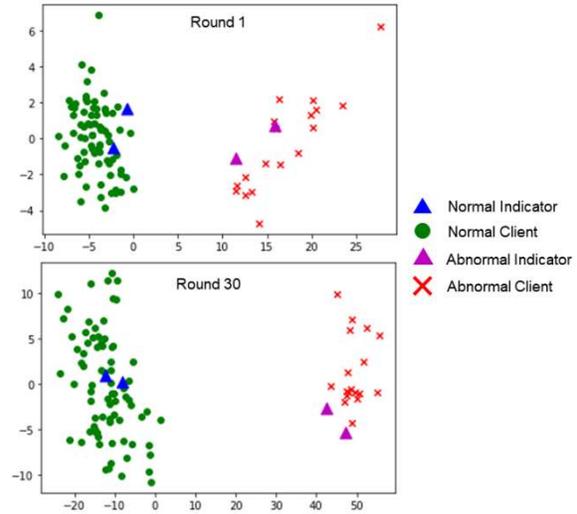

**FIGURE 3.** Projected vectors on 2-d space after reducing the dimension of vectors by PDA

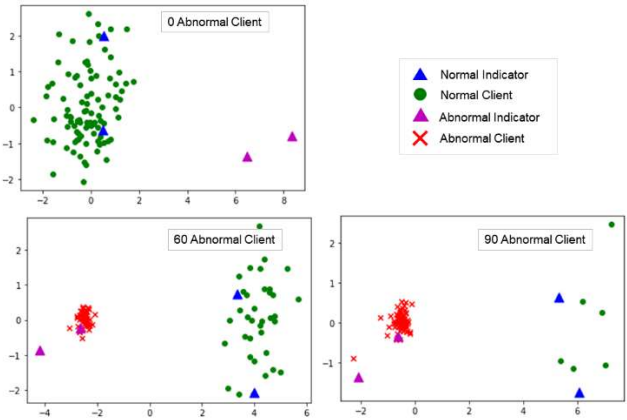

**FIGURE 4.** Projected vectors on 2-d space in different numbers of abnormal client conditions.

## V. CONCLUSION

Federated Learning is an outstanding method for the protection of private and personal data. However, it opens the door for malevolent locals. This paper indicates another supervised method for protecting the global model from malicious locals by utilizing KNN clustering. The proposed algorithm can diminish the effects of Sybils. And it is also enabled to be applied to any training round.

## E. REFERENCES